\documentclass[10pt,twocolumn,letterpaper]{article}

\usepackage[pagenumbers]{cvpr}            %

\usepackage{catchfile} %
\usepackage{adjustbox}
\usepackage{multirow}
\usepackage{colortbl}

\usepackage[accsupp]{axessibility} %

\usepackage{lipsum}  
\usepackage{array}
\usepackage{makecell}
\usepackage{soul}
\newcolumntype{H}{>{\setbox0=\hbox\bgroup}c<{\egroup}@{}}
\usepackage{color}
\usepackage{stfloats}
\definecolor{light}{rgb}{0.5, 0.5, 0.5}

\usepackage{amsmath}
\DeclareMathOperator*{\argmax}{arg\,max}

\newcommand{\greenup}{\textcolor{ForestGreen}{$\uparrow$}}
\newcommand{\reddown}{\textcolor{BrickRed}{$\downarrow$}}
\definecolor{applegreen}{rgb}{0.05, 0.60, 0.20}

\newcommand{\PAR}[1]{\vskip8pt \noindent {\bf #1~}}
\newcommand{\PARbegin}[1]{\noindent {\bf #1~}}

\definecolor{cvprblue}{rgb}{0.21,0.49,0.74}
\definecolor{brickred}{rgb}{0.8, 0.25, 0.33}
\usepackage[pagebackref,breaklinks,colorlinks,allcolors=cvprblue]{hyperref}
\usepackage{soul}
\def\paperID{26} %
\def\confName{CVPR}
\def\confYear{2025}

\title{What is the Added Value of UDA in the VFM Era?}

\author{Brunó B. Englert \quad Tommie Kerssies \quad Gijs Dubbelman \\
Eindhoven University of Technology \\
{\tt\small \{b.b.englert, t.kerssies, g.dubbelman\}@tue.nl}
}

\begin{document}
\maketitle
\begin{abstract}
\textbf{Unsupervised Domain Adaptation (UDA)} can improve a perception model's generalization to an unlabeled target domain starting from a labeled source domain. UDA using \textbf{Vision Foundation Models (VFMs)} with synthetic source data can achieve generalization performance comparable to fully-supervised learning with real target data. However, because VFMs have strong generalization from their pre-training, more straightforward, source-only fine-tuning can also perform well on the target. As data scenarios used in academic research are not necessarily representative for real-world applications, it is currently unclear (a) how UDA behaves with more representative and diverse data and (b) if source-only fine-tuning of VFMs can perform equally well in these scenarios. Our research aims to close these gaps and, similar to previous studies, we focus on semantic segmentation as a representative perception task. We assess UDA for synth-to-real and real-to-real use cases with different source and target data combinations. We also investigate the effect of using a small amount of labeled target data in UDA. We clarify that while these scenarios are more realistic, they are not necessarily more challenging. Our results show that, when using stronger synthetic source data, UDA's improvement over source-only fine-tuning of VFMs reduces from +8 mIoU to +2 mIoU, and when using more diverse real source data, UDA has no added value. However, UDA generalization is always higher in all synthetic data scenarios than source-only fine-tuning and, when including only 1/16 of Cityscapes labels, synthetic UDA obtains the same state-of-the-art segmentation quality of 85 mIoU as a fully-supervised model using all labels. Considering the mixed results, we discuss how UDA can best support \textbf{robust autonomous driving at scale}.
\end{abstract}
    
\section{Introduction}
\label{sec:intro}
Perception of the environment through computer vision is a key research topic in pursuit of \textbf{Autonomous Driving} (AD). Whether it is as part of an end-to-end~\cite{chen2023e2esurvey} or modular system, vision models provide an autonomous vehicle with the crucial ability to perceive and understand its surroundings. Developing vision models that are robust to diverse data distributions, including previously unseen distributions, is key to safe autonomous driving and a fundamental challenge in machine learning. This challenge becomes particularly pronounced when vision models encounter significant distribution shifts between the training and testing conditions. In dense perception tasks like instance-, semantic-, and panoptic segmentation, obtaining accurate pixel-level labeled data is labor-intensive, inherently limiting data scale and diversity, which can hinder robustness to real-world conditions.

\begin{figure}[t]
    \centering
    \includegraphics[width=\columnwidth]{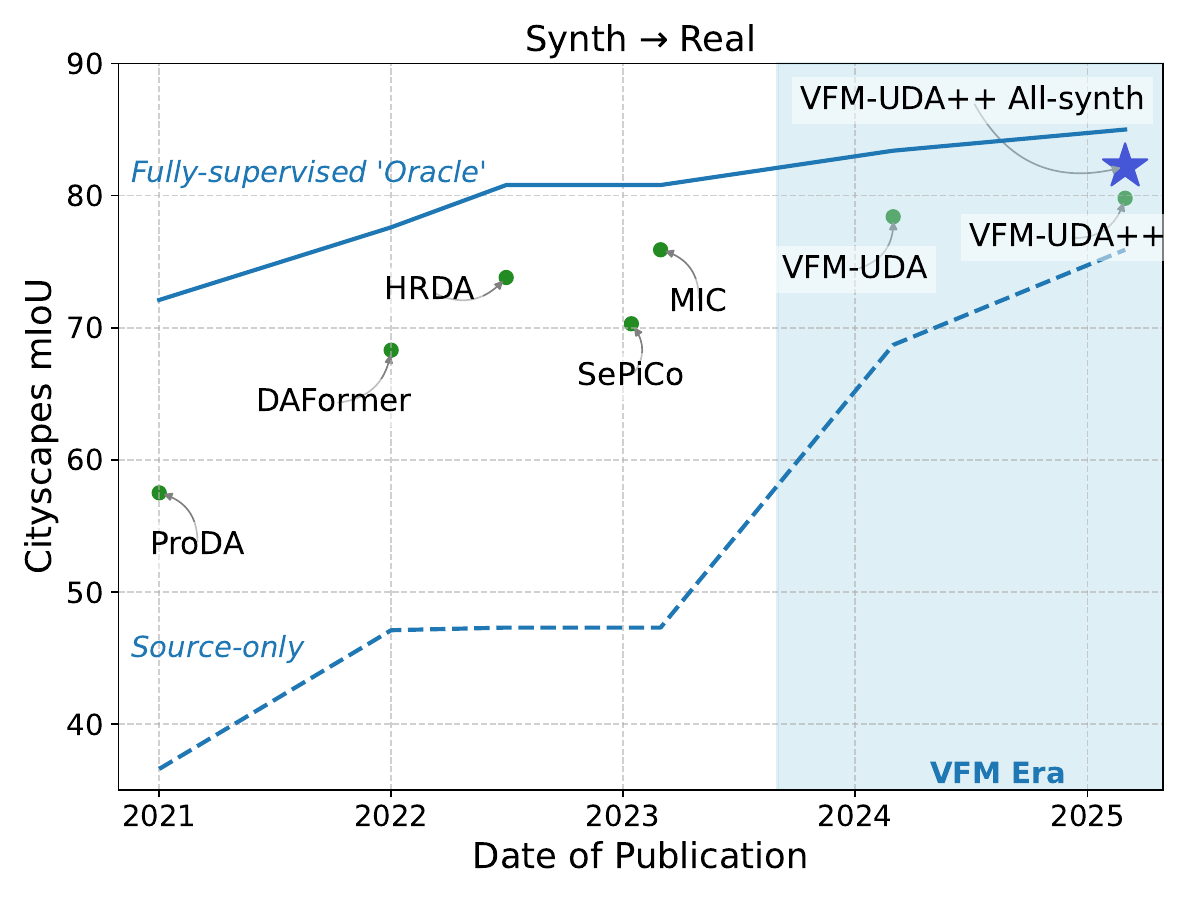}
    \caption{\label{fig:eye-catcher} \textbf{UDA methods vs. source-only baselines and fully-supervised oracles.} While VFM-based UDA achieves generalization close to fully-supervised learning, the added value of UDA over simple source-only fine-tuning requires further investigation.}
\end{figure}

\textbf{Unsupervised Domain Adaptation} (UDA)~\cite{Hoffman_featalign_16} addresses distribution shifts, often called \textit{domain gaps}, by adapting models trained on a labeled \textit{source domain} and improving their generalization to an unlabeled \textit{target domain}. Without adaptation, variations in location, weather, lighting, textures, sensor properties, and class distributions can lead to overfitting on the source domain and poor generalization to the \textit{target domain}. A key benefit of UDA over fully-supervised learning is that it can leverage unlabeled data to improve generalization, as obtaining and using unlabeled data is significantly more scalable than labeled data. Recently, UDA methods leveraging \textbf{Vision Foundation Models} (VFMs) with diverse, labeled synthetic source data and unlabeled real target data, have reached levels of generalization very close to fully-supervised learning on the target data~\cite{vfm-uda++}. These strong results of state-of-the-art VFM-based UDA can be observed in \cref{fig:eye-catcher}. Thus, potentially, UDA allows training a model for an unlabeled \textit{target domain} that is practically as effective as a fully-supervised model that requires many labels for the \textit{target domain}. However, also shown in \cref{fig:eye-catcher}, the performance of source-only fine-tuning has caught up with UDA through the use of VFMs. This makes us question the utility of UDA in the VFM era. 

The potential \textbf{utility of UDA in AD} is in two prominent use cases. In the first \textit{synth-to-real} use case, the goal is to reduce the need for real labeled data by using synthetic, automatically labeled data as much as possible. In this use case, UDA is applied from a labeled synthetic source domain to an unlabeled real target domain, to improve the perception model's general ability to perform well on real data. This use case is particularly relevant as it not only reduces labeling costs but also enables the synthetic generation of rare edge cases that are otherwise difficult to collect. As such, training with synthetic data can improve robustness to rare and unforeseen circumstances. In the second \textit{real-to-real} use case, the goal is to improve the perception model's quality such that it also performs well for a domain for which no labels are available. In this use case, UDA is applied from a labeled real source domain to an unlabeled real target domain, to transfer the knowledge from this labeled source domain to the unlabeled target domain. This use case is potentially relevant when an autonomous vehicle needs to be deployed in a new location or condition, without the possibility of retraining its perception models beforehand on newly collected and labeled data. Both use cases have been extensively researched in an academic setting but with relatively limited benchmarks.

Since the start of UDA research almost a decade ago, more synthetic and real data has become available, likely even more so in industry than academia. Furthermore, with the economic relevance of autonomous driving, it can be assumed that a small amount of labeled data can be made available for practically any AD domain. However, collecting enough real edge-case examples remains a challenge. Given these factors and the rise of VFMs, the role of UDA in perception must be re-evaluated for realistic autonomous driving scenarios.

Therefore, in this study, \textbf{we systematically evaluate VFM-based UDA} for synth-to-real and real-to-real adaptation scenarios, diversifying both source and target data. Additionally, we investigate whether UDA retains its added value when small amounts of labeled target data are available. In~\cref{sec:preliminary}, we outline the training settings and methodologies used in our experiments. We compare UDA with source-only fine-tuning across multiple data scenarios, described in~\cref{sec:scenarios}, to assess its practical utility. Results for synth-to-real and real-to-real adaptation are presented in~\cref{sec:synth_results} and~\cref{sec:real_results}, respectively. In~\cref{sec:discussion} and~\cref{sec:conclusions}, we discuss the implications of our findings and conclude that UDA’s primary benefit in autonomous driving (AD) lies in synth-to-real adaptation. Specifically, it provides an additional safeguard against potential limitations in the diversity of labeled synthetic source data that could otherwise impact perception quality in an unseen real-world target domain.

\textbf{Our contribution} is to systematically examine whether and how UDA provides added value in the VFM era, focussing on realistic autonomous driving scenarios. We provide new insights into UDA’s efficacy and limitations in AD, which has always been one of its key potential applications, as witnessed by the many related academic benchmarks. We hope our findings guide future UDA work toward more representative data scenarios and practical AD applications.

\section{Related Work}
\label{sec:relwork}

\begin{table}[tb] 
\centering 
\begin{adjustbox}{width=\linewidth}
\begin{tabular}{lcccccc} 
\toprule
Setting & Architecture & Encoder & Source & Target  & \thead{CS \\ mIoU} \\
\toprule

UDA & SePiCo~\cite{xie_sepico_2022} & MiTB-5 & Synthia & CS & 66.5 \\
UDA & MIC~\cite{mic_2023} & MiT-B5 & Synthia & CS & 67.3 \\
UDA & HRDA~\cite{hoyer_hrda_2022}~+~PiPa~\cite{Chen_pipa_2023}  & MiTB-5 & Synthia & CS  & 68.2 \\
UDA & HRDA~\cite{hoyer_hrda_2022}~+~CDAC~\cite{Wang2023CDACCA} & MiTB-5 & Synthia & CS  & 68.7 \\
\rowcolor{gray!30}
UDA & VFM-UDA++~\cite{vfm-uda++} & DINOv2-B &  Synthia & CS & \textbf{69.7}  \\

\midrule

UDA & HRDA~\cite{hoyer_hrda_2022}~+~DiGA~\cite{shen2023diga} & MiTB-5   & GTA5 & CS  & 74.3 \\
UDA & HRDA~\cite{hoyer_hrda_2022}~+~CDAC~\cite{Wang2023CDACCA} & MiTB-5 & GTA5 & CS         & 75.3 \\
UDA & Diffusion~\cite{PengDiffusion}                         & MiTB-5   & GTA5 & CS        & 75.3 \\
UDA & HRDA~\cite{hoyer_hrda_2022}~+~PiPa~\cite{Chen_pipa_2023} & MiTB-5 & GTA5 & CS         & 75.6 \\
UDA & MIC~\cite{mic_2023}                                    & MiTB-5   & GTA5 & CS       & 75.9 \\
UDA & VFM-UDA~\cite{englert2024exploring}          & DINOv2-B           & GTA5 & CS   & 77.1 \\
UDA & DCF~\cite{chen2024transferring} \textbackslash w depth & MiTB-5   & GTA5 & CS         & 77.7 \\
\rowcolor{gray!30}
UDA & VFM-UDA++~\cite{vfm-uda++}  & DINOv2-B  & GTA5 & CS      & \textbf{79.1}  \\

\midrule

Source-only  & Rein~\cite{wei2024stronger}      & DINOv2-L & GTA5, S, US & - & 78.4 \\
\rowcolor{gray!30}
Source-only  & VFM-UDA++~\cite{vfm-uda++} & DINOv2-L & GTA5, S, US & - & \textbf{79.7} \\

\midrule

Semi-sup.  & AllSpark~\cite{allspark}     & MiT-B5 & 1/16 CS & CS & 78.3 \\
Semi-sup.  & BeyondPixels~\cite{howlader2024beyond} & RN-101 & 1/16 CS & CS & 78.5 \\
Semi-sup.  & SemiVL~\cite{hoyer2024semivl}       &  CLIP-B & 1/16 CS & CS & 77.9 \\
\rowcolor{gray!30}
Semi-sup.  & VFM-UDA++~\cite{vfm-uda++}    & DINOv2-L & 1/16 CS & CS & 83.5 \\
Semi-sup.  & UniMatchV2~\cite{unimatchv2}   & DINOv2-B  & 1/16 CS & CS & \textbf{83.6} \\

\midrule

\rowcolor{gray!30}
Fully-sup & VFM-UDA++~\cite{vfm-uda++} & DINOv2-L  & CS              & - & 85.0 \\
Fully-sup & ViT-Adapt~\cite{chen2022vitadapter}~+~M2F~\cite{cheng2021mask2former}     & DINOv2-L  & CS              & - & 84.9 \\
Fully-sup & MetaPrompt~\cite{wan2023harnessing} & StableDiff1.5  & CS              & - & \textbf{86.0} \\

\bottomrule
\end{tabular}
\end{adjustbox} 
\caption{\label{tab:vfm-uda++} \textbf{Comparison of VFM-UDA++~\cite{vfm-uda++} with state-of-the-art methods evaluated on Cityscapes~\cite{cordts_cityscapes_2016}.} Despite being designed for UDA, VFM-UDA++ performs competitively across settings, making it a strong choice for evaluating UDA’s added value.}
\end{table}

\PARbegin{Vision Foundation Models (VFMs).}
VFMs have emerged as powerful tools, significantly advancing performance across various computer vision tasks through extensive pre-training on large-scale datasets. Models such as CLIP~\cite{radford_clip_2021}, MAE~\cite{mae_2022}, EVA-02~\cite{eva02_2023}, and DINOv2~\cite{dinov2_2023} provide robust representations that generalize exceptionally well across tasks and domains, even with limited labeled data. DINOv2, in particular, has demonstrated superior performance in dense prediction and domain generalization tasks \cite{wei2024stronger, englert2024exploring, kerssies2024benchmarking, vu2024bravo, kerssies2024evaluating, kerssies2025eomt}, substantially outperforming traditional ImageNet~\cite{deng2009imagenet} pre-trained architectures. Recent studies of label-efficient semantic segmentation use VFMs for their robust generalization capabilities. In the following paragraphs, we outline these learning strategies.\\

\PAR{Domain Generalized Semantic Segmentation (DGSS).}
DGSS targets generalizing segmentation models to unseen domains without accessing \textit{target domain} data during training. Earlier methods relied primarily on data augmentation~\cite{advstyle, PASTA, gtrltr} and domain-invariant feature extraction techniques~\cite{ibn, choi2021robustnet, dg:tang2020selfnorm}. Recent VFM-based DGSS approaches, such as Rein~\cite{wei2024stronger}, use parameter-efficient fine-tuning to preserve the knowledge learned in pre-training.\\

\PAR{Semi-supervised Semantic Segmentation (SSS).}
SSS seeks to minimize annotation costs by combining limited labeled data with unlabeled data from the same domain. Early SSS methods used offline pseudo-labeling, which was computationally expensive~\cite{st++, simplebaseline}. More recent techniques, including FixMatch~\cite{fixmatch} and UniMatch~\cite{unimatch}, adopted efficient online pseudo-labeling with weak-to-strong consistency regularization. Recent incorporation of VFMs into semi-supervised methods, shown by UniMatchV2~\cite{unimatchv2}, further enhanced generalization and simplified the reliance on specialized semi-supervised components due to their inherent robust representations.\\

\PAR{Unsupervised Domain Adaptation (UDA).}
UDA addresses the challenge of generalizing models trained on labeled source data to perform well in an unlabeled target domain under substantial domain shifts, similar in objective to Domain Generalized Semantic Segmentation (DGSS). However, unlike DGSS, UDA explicitly leverages unlabeled data from the target domain during training, often focusing on bridging the gap between easy-to-generate synthetic source data and easy-to-collect real target data. Its approach shares strong similarities with semi-supervised semantic segmentation (SSS), as both employ unlabeled data and pseudo-labeling strategies. However, the key difference is the presence of a domain gap between the source and target data in UDA. UDA methods~\cite{laine2017temporal, araslanov_self-supervised_2021, cluda_2022, xie_sepico_2022,  tranheden_dacs_2021, romijnders2019domainnormlayer} rely on pseudo-labeling, consistency regularization~\cite{ chen2024transferring}, adversarial training~\cite{Hoffman_cycada_2018}, and entropy minimization~\cite{vu2018advent}. Recent frameworks like DAFormer~\cite{hoyer_daformer_2022} and HRDA~\cite{hoyer_hrda_2022} incorporated transformer-based multi-scale architectures, significantly improving generalization capabilities. Techniques like Masked Image Consistency (MIC)~\cite{mic_2023} further enhanced spatial consistency. The introduction of VFM-based UDA, such as VFM-UDA~\cite{englert2024exploring} and its improved variant VFM-UDA++~\cite{vfm-uda++}, demonstrated remarkable generalization gains by leveraging powerful single-scale VFMs such as DINOv2 combined with multi-scale inductive biases and VFM-based feature distillation losses.\\

\PAR{UDA compared to other settings.} In this work, we investigate the continued relevance of UDA by comparing it to simple source-only fine-tuning across different realistic data scenarios. Rather than evaluating all possible UDA approaches, we select a strong representative method to ensure meaningful comparisons. As shown in Table~\ref{tab:vfm-uda++}, the recent VFM-UDA++~\cite{vfm-uda++} achieves state-of-the-art results on standard UDA benchmarks such as Synthia~\cite{ros2016synthia} to Cityscapes~\cite{cordts_cityscapes_2016} and GTA5~\cite{richter_playing_nodate} to Cityscapes while also performing competitively in other settings. For source-only fine-tuning, VFM-UDA++ surpasses the previous state-of-the-art method, Rein~\cite{wei2024stronger}, without requiring parameter-efficient fine-tuning. Similarly, in semi-supervised fine-tuning, it remains competitive with the recent method UniMatchV2~\cite{unimatchv2}, despite not being explicitly designed for SSS. In fully-supervised fine-tuning, it performs within 1 mIoU of the best available method. Given these results, we use VFM-UDA++ as our representative UDA method to assess the added value of UDA across different scenarios.

\section{Preliminaries} 
\label{sec:preliminary}

In~\cref{sec:training_settings} we first describe the training settings used in this work that do not involve domain adaptation. Next, in~\cref{sec:naive_uda} we outline the fundamental formulation of naive Unsupervised Domain Adaptation (UDA), which serves as the conceptual foundation for modern state-of-the-art UDA approaches discussed in~\cref{sec:uda}. Finally, in~\cref{sec:two-stage}, we introduce a UDA approach with a small amount of target labels. The methods presented here form the basis of our experimental setup; rather than proposing new methods, we evaluate the effectiveness of existing approaches in practical scenarios.

\subsection{Training settings without adaptation}
\label{sec:training_settings}

\PARbegin{Source-only.}
Source-only fine-tuning represents the simplest setting, equivalent to the domain-generalized semantic segmentation (DGSS) setting. In this setting, the model is fine-tuned exclusively on labeled source domain data, with no access to target domain data during training. The key objective is to assess the model’s ability to generalize to unseen target domains without being specifically adapted to them during fine-tuning.

\PAR{Semi-supervised.}
The semi-supervised setting assumes no significant domain gap between source and target domains. Here, both labeled and unlabeled data are drawn from the same domain, where only a limited amount of data is labeled. The training pipeline matches exactly the UDA training pipeline, as discussed in~\cref{sec:uda}.

\PAR{Fully-supervised.}
In the fully-supervised setting, the model is trained with access to labeled data from the target domain, allowing it to learn directly from the distribution it is evaluated on. This setting provides an upper bound on performance, as it eliminates the challenges associated with domain shift and adaptation, serving as a reference point for evaluating the effectiveness of other learning strategies.

\subsection{Naive UDA}
\label{sec:naive_uda}

Let $f_\theta$ denote a neural network parameterized by $\theta$ that maps an image $x \in \mathbb{R}^{H \times W \times 3}$ to a semantic segmentation prediction $\hat{y} \in \mathbb{R}^{H \times W \times K}$, where $K$ is the number of classes.

Given a labeled source domain dataset $\mathcal{D}^S = \{(x_m^S, y_m^S)\}_{m=1}^{N_S}$ and $\mathcal{D}^T = \{x_n^T\}_{n=1}^{N_T}$ an unlabeled target domain dataset, naive UDA follows a three-stage process:

\begin{enumerate}
    \item \textbf{Source-domain training.} A model with parameters $\theta_S$ is first trained on the source-domain data by minimizing a vanilla cross-entropy loss:

\begin{equation}
    \mathcal{L}(\theta, \mathcal{D}^{S}) = \mathcal{H}(f_{\theta}(x^S), y^S).
\end{equation}

    \item \textbf{Pseudo label generation.} The trained source model $f_{\theta_S}$ is frozen and used to generate pseudo-labels $y^{pseudo}$ for the unlabeled target domain data:
\begin{equation}
    y_{nijk}^{pseudo} = [k = \argmax_{k'} f_{\theta_S}(x_n^T)_{ijk'}]\,,
\end{equation}
where $[\cdot]$ denotes the Iverson bracket selecting the most confident class prediction.

    \item \textbf{Target domain adaptation.} 
    The pseudo-labels $y^{pseudo}$ are then used to supervise the training of a \textbf{new model} with parameters $\theta_T$ on both the source and pseudo-labeled target domains, leading to the combined loss:

\begin{equation}
    \mathcal{L}^{\text{total}}(\theta_T) = \mathcal{L}(\theta_S, \mathcal{D}^S) + \mathcal{L}(\theta_T, \mathcal{D}^{T}_{pseudo}).
\end{equation}

\end{enumerate}

Although straightforward, this naive UDA approach suffers from limitations such as error accumulation, motivating the development of more sophisticated adaptation techniques discussed in the next section.

\subsection{Modern UDA}
\label{sec:uda}

Modern UDA methods mitigate the limitations of naive pseudo-labeling by employing a single-stage learning framework, where the pseudo labels are continuously updated during training rather than using a separate pseudo-label generation phase. These methods introduce mechanisms to reject uncertain pseudo-labels, enhance feature alignment, and improve spatial consistency, in an online manner, leading to a higher performing model.

Most UDA methods utilize an EMA~\cite{tarvainen_mean_2017} teacher network $f_{\theta^{teacher} }$, parameterized by $\theta^{teacher}$, updated as:
\begin{equation}
    \theta_{t+1}^{teacher} \leftarrow \alpha \theta_{t}^{teacher} + (1 - \alpha)\theta_{t+1}^{student}\,,
\end{equation}
where $t$ denotes the training iteration and $\alpha$ denotes the momentum factor controlling the update rate. The EMA teacher provides the pseudo-labels, weighted by a quality estimate $q$~\cite{french_semi-supervised_2020, hoyer_daformer_2022}, which quantifies confidence in the pseudo-labels, to reduce uncertainty:
\begin{equation}
    q^T = \frac{\sum_{i=1}^{H}\sum_{j=1}^{W}[\max_{k'}f_{\theta^{teacher}}(x^T)_{ijk'} > \tau]}{H \cdot W}\,,
\end{equation}
with threshold $\tau$. We use this estimate $q$ to weigh the cross entropy loss of the target domain.  

Furthermore, state-of-the-art methods utilize a feature distance (FD) loss to align features between the source and target domains~\cite{hoyer_daformer_2022, hoyer_hrda_2022, mic_2023, vfm-uda++}. Specifically, features extracted from a pre-trained model are used to compute a distance-based regularization term between the pre-trained model and the student model. This forces the student model to not forget its large-scale pre-training, thereby enhancing adaptation performance. The implementation of FD loss depends on the feature extractor, the chosen distance metric, and the specific network layers used for alignment.

Finally, Masked Image Consistency (MIC)~\cite{mic_2023} ensures that predictions remain consistent when input patches are masked, encouraging robustness to missing information. For a given mask $\mathcal{M}$ and masked target image $x^{M}=\mathcal{M}\odot x^T$, MIC enforces consistency between masked predictions and pseudo-labels predicted from non-masked images:
\begin{equation}
    \mathcal{L}^{M}(\theta) = \mathcal{H}(f_{\theta}(x^{M}), p^{T})\,.
\end{equation}
This forces the model to reason based on context and predict consistent segmentation maps for the masked regions. 

The final training objective combines all terms:
\begin{equation} 
\begin{split}
    \mathcal{L}^{\text{total}}(\theta) & = \mathcal{L}^{S}(\theta) + q^T\lambda^{T}\mathcal{L}^{T}(\theta) \\ 
    & \quad + q^T \lambda^{M} \mathcal{L}^{M}(\theta) + \lambda^{FD}\mathcal{L}^{FD}(\theta)\,,
\end{split}
\end{equation}
where $\lambda^{T}$, $\lambda^{M}$, and $\lambda^{FD}$ are weighting factors that control the relative influence of each loss term.

\subsection{UDA with few target labels}
\label{sec:two-stage}
In the context of UDA, it is unclear how to optimally integrate labeled target domain data into training. Inspired by two-stage fine-tuning methods from supervised learning, we adopt a similar approach in UDA, involving two distinct training stages:
\begin{itemize}
    \item \textbf{Stage 1: UDA with arbitrary large-scale data.} Initially, we employ standard UDA using a large-scale labeled source dataset (e.g., synthetic data) and an unlabeled real-world target dataset to establish strong generalization.
    \item \textbf{Stage 2: Semi-supervised learning on target domain.} Subsequently, we reuse the same UDA pipeline, now in a semi-supervised fashion, using labeled and unlabeled data from the same target domain. Specifically, a small amount of labeled target data is combined with a larger set of unlabeled target data, refining the model’s performance specifically for the domain of interest.
\end{itemize}

\section{Experiments}
\label{sec:results}
Our experiments assess the added value of UDA over source-only fine-tuning in data scenarios relevant to autonomous driving. As is common in UDA works, we use semantic segmentation as a representative perception task. It involves dense multi-class predictions that require fine-grained and contextual image understanding.

\subsection{Implementation details}
We mostly follow the same implementation as VFM-UDA++~\cite{vfm-uda++}, using a DINOv2-L~\cite{dinov2_2023} encoder with a ViT-Adapter~\cite{chen2022vitadapter} and a BasicPyramid~\cite{vfm-uda++} decoder. Training runs for 40,000 iterations with AdamW~\cite{loshchilov2018adamw}, a batch size of 8, and learning rates of 1.4$\times10^{-4}$ for the decoder and 1.4$\times10^{-5}$ for the encoder, with 0.9 layerwise decay.

\begin{figure*}
    \centering
    \includegraphics[width=1\linewidth]{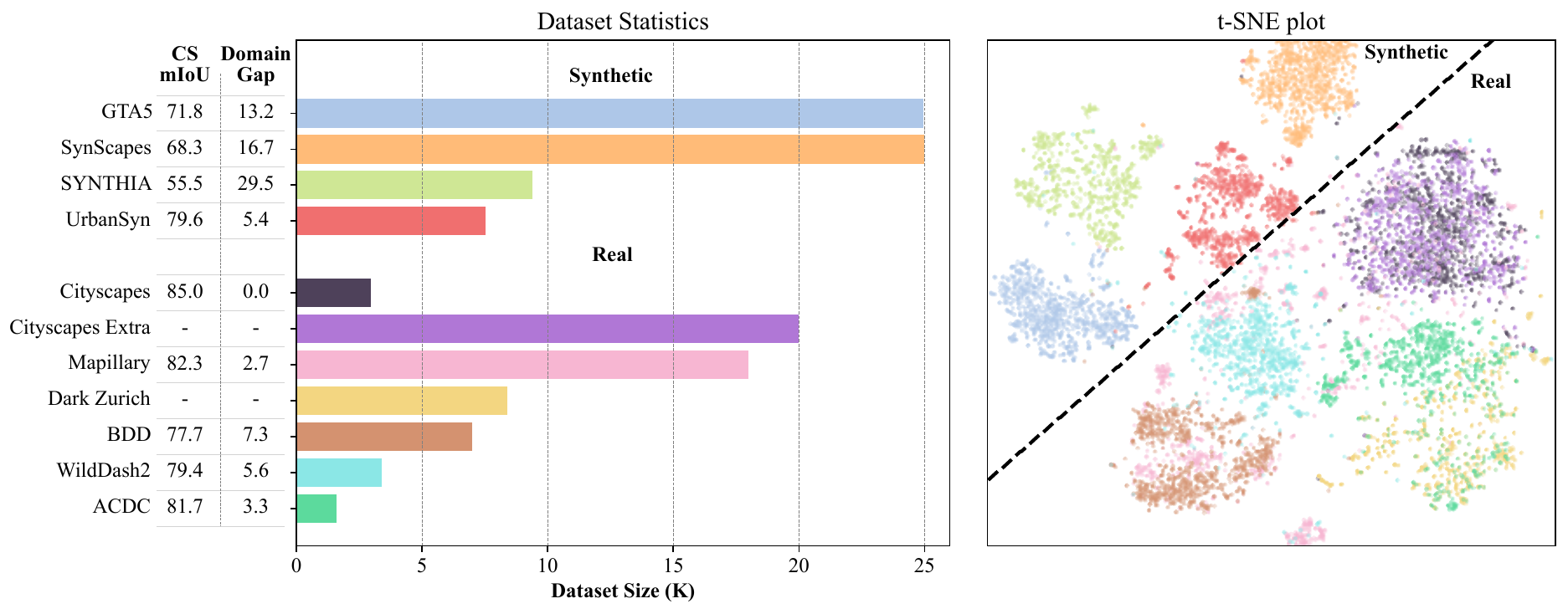}
    \caption{\label{fig:infographic}\textbf{Dataset Overview.}
    On the top left, we show source-only performance (mIoU) evaluated on Cityscapes~\cite{cordts_cityscapes_2016} for models trained on different source datasets, including their domain gaps relative to the oracle (trained on Cityscapes), and dataset sizes (in thousands of samples). On the top right, a t-SNE visualization of DINOv2-L~\cite{dinov2_2023} \texttt{[CLS]} token embeddings shows a clear separation between synthetic and real datasets, while also capturing semantic similarities among them.}
\end{figure*}

\subsection{Data scenarios and datasets} 
\label{sec:scenarios}
We use a combination of commonly available and often used academic benchmark datasets to support reproducing our results and the ability to cross-compare our results with those in earlier and future work. Continuing on~\cref{sec:intro}, we envision the following use cases of UDA as the basis for our experimental setup.

\PAR{UDA use cases.} The first \textit{synth-to-real} UDA use case aims to reduce the dependency on labeled real data by using automatically labeled synthetic data. Here we assess the importance of scaling and diversifying labeled synthetic data, scaling and diversifying unlabeled real data, and the impact of introducing small amounts of labeled real data. In this use case, measuring how variations in synthetic and real data influence UDA and its main competitor, source-only fine-tuning, is our priority. Specifically, we experiment with the following data scenarios:
\begin{itemize}
    \item \textbf{Synth 1: Scaling source data.} In our first data scenario, we scale and diversify synthetic source data to measure the effect on the generalization to the target for source-only fine-tuning, naive UDA, and UDA.
    \item \textbf{Synth 2: Scaling source and target data.} In this scenario, we measure the effect of scaling and diversifying target data for different source data compositions.
    \item \textbf{Synth 3: Mixing in few target labels.} Here, we measure the effect of mixing in a small amount of labeled real target data for synth-to-real UDA.
\end{itemize}

For the second \textit{real-to-real} UDA use case, we assess the added value of UDA when diverse real source data is available and the goal is to obtain a vision model for a specific target domain. Here we test if the UDA model can specialize or improve for the target domain compared to source-only fine-tuning on the diverse source data. Also, in this scenario, we introduce a small amount of labeled target data. Specifically, we experiment with the following data scenarios:
\begin{itemize}
    \item \textbf{Real 1: Scaling source data.} In our first data scenario, we assess the effect of using more diverse real source data, again comparing source-only fine-tuning and UDA.
    \item \textbf{Real 2: Mixing in few target labels.} Also for the real-to-real use case, we measure the effect of mixing in a small amount of labeled target data.
\end{itemize}

One can argue that our data scenarios make the task of UDA easier. Because we include more diverse synthetic and real source data and even include target labels, we make the domain gap between source and target domain significantly easier to bridge. This is arguably true, but we believe this resembles most use cases of UDA in autonomous driving. Diverse synthetic and real source domain datasets are simply available, and probably even better datasets are available in industry, covering many scenes, objects, and challenging environmental conditions. Thus, only a challenging target-domain dataset could increase the domain gap. However, we believe that for most applications of UDA one would want to record data at one's convenience, and not having to wait for a specific moment when specific challenging environmental conditions, \eg, snow, hail, fog, are met. One would want the VFM to learn diverse scenes, objects, and challenging conditions from pre-training and rich source data, and transfer this knowledge to the target domain via UDA, without requiring the same variations to be recorded in the target data.
Therefore, it is very important to not only measure UDA performance on the target dataset but also on a challenging out-of-target dataset that is not used during UDA~\cite{Piva_genstdy_23,englert2024exploring, vfm-uda++}. This allows us to measure how much general knowledge from pre-training and the source data is preserved after UDA. This is very important and often not considered in UDA research, but we will use this validation approach, as is explained next.

\PAR{Target datasets.} The Cityscapes (CS) dataset~\cite{cordts_cityscapes_2016} serves as the unlabeled target domain for the scenarios above, as is common in most UDA works. Arguably, Cityscapes represents a relatively narrow target domain compared to, \eg, Mapillary Vistas~\cite{Neuhold_mapillary_17} or ACDC~\cite{Sakaridis2021acdc}. As just explained, we do this purposefully to examine how different learning strategies can specialize to a focused target domain without `forgetting' general knowledge. To measure forgetting, we use the challenging WildDash2 (WD2) dataset~\cite{Zendel_wilddash_18}, which consists of many diverse scenes and environmental conditions. The goal of UDA is not to improve on WildDash2 per se, but to maintain comparable mIoU by preserving general knowledge from VFM pre-training and the source data. When scaling unlabeled target data, we either use Cityscapes-extra to add more data from the same distribution or diversify the target domain by extending it with BDD~\cite{yu2020bdd100k}, Mapillary Vistas, Dark Zurich~\cite{dai2018darkzurich}, and ACDC.

\PAR{Source datasets.} For synth-to-real scenarios, UDA research often uses GTA5~\cite{richter_playing_nodate} as a single labeled source dataset. This study extends it with different combinations of SYNTHIA~\cite{ros2016synthia}, UrbanSyn~\cite{gomez2023urbansyn}, and SynScapes~\cite{Wrenninge2018Synscapes}, when scaling synthetic labeled source data. 
For the real-to-real scenarios, we use BDD~\cite{yu2020bdd100k} as a single labeled source dataset, and extend it with Mapillary Vistas~\cite{Neuhold_mapillary_17} and ACDC~\cite{Sakaridis2021acdc}, when scaling real labeled source data.
These dataset combinations scale the data available to UDA, but more importantly diversify the information provided to UDA. In \cref{fig:infographic}, statistics of the used source and target datasets are provided, including dataset size, their domain gap to Cityscapes, a t-SNE plot of their DINOv2~\cite{dinov2_2023} embeddings, and visual examples.

\PAR{Few target labels.} To complete our experiments, we also validate scenarios where a small amount of labeled target data is available. We do this by using 186 labeled Cityscapes images of the commonly used 1/16 Cityscapes split~\cite{hoyer2024semivl}. This is a relevant scenario, as the economic value of autonomous driving allows investing in labeling a small amount of data for many relevant domains. By comparing UDA with few Cityscapes labels to fully-supervised learning using all labeled Cityscapes images, we can better assess the utility of UDA in scenarios where a small amount of labeled data is available for the target domain.

\subsection{Synth-to-real results}
\label{sec:synth_results}

\PARbegin{Scenario Synth 1: Scaling source data.}
In the first scenario, the focus is on scaling and diversifying labeled synthetic source data. We compare the UDA method VFM-UDA++ against naive UDA (see~\cref{sec:naive_uda}) and source-only fine-tuning. All approaches use the same architecture as VFM-UDA++ to allow for a fair, insightful comparison. The results are provided in~\cref{tab:scenario1} with the upper part using only GTA5 as source and the bottom part using GTA5, SYNTHIA, and UrbanSyn. First, naive UDA performs significantly worse than UDA in both settings, confirming that a naive approach to UDA is ineffective. Second, by comparing the numbers, we can observe that the added value of UDA over source-only fine-tuning reduces from +8.0 mIoU to +1.8 mIoU when using more synthetic source datasets. This is simply because source-only fine-tuning with all synthetic data has a significantly higher starting point. It is much harder to improve on this as it gets closer to the upper bound of 85 mIoU of fully-supervised learning. Finally, the source-only models provide the references for performance on WildDash2, and we can observe that UDA does not `forget' general knowledge contained in the source domain by adapting it to the target. It can even improve on it for this synth-to-real scenario, as WildDash2 is also a real dataset unlike the synthetic source datasets. \textbf{To conclude, when scaling synthetic source data, UDA performance increases, but its added value over source-only fine-tuning diminishes.}

\begin{table}[h] 
\centering 
\begin{adjustbox}{width=0.95\linewidth}
\begin{tabular}{lcccc} 
\toprule
Setting& Source & Target & mIoU CS & mIoU WD2 \\
\toprule
\rowcolor{gray!30}
Source-only FT  & GTA5       & -     & 71.8\phantom{..↑.+2.2}                               & 65.5\phantom{.}\phantom{.↑.+1.4}  \\
Naive UDA       & GTA5       & CS    & 73.6\phantom{.}\greenup~\textcolor{applegreen}{+2.2} & 66.9\phantom{.}\greenup~\textcolor{applegreen}{+1.4} \\
UDA             & GTA5       & CS    & 79.8\phantom{.}\greenup~\textcolor{applegreen}{+8.0} & 69.0\phantom{.}\greenup~\textcolor{applegreen}{+3.5} \\
\midrule
\rowcolor{gray!30}
Source-only FT  & GTA5, S, US & -   & 79.7\phantom{sas+1.0} & 67.3\phantom{sas+1.6} \\
Naive UDA       & GTA5, S, US & CS  & 80.7~\greenup~\textcolor{applegreen}{+1.0} & 68.9~\greenup~\textcolor{applegreen}{+1.6} \\
UDA             & GTA5, S, US & CS  & 81.5~\greenup~\textcolor{applegreen}{+1.8} & 70.7~\greenup~\textcolor{applegreen}{+3.4}  \\
\bottomrule
\end{tabular}
\end{adjustbox} 
\caption{\label{tab:scenario1} \textbf{Scenario Synth 1: Scaling source data.} When increasing labeled synthetic source data, the added value of UDA over source-only fine-tuning reduces significantly. 
} 
\end{table}

\PAR{Scenario Synth 2: Scaling source and target data.}
In this scenario, we scale the amount of unlabeled target data under both less diverse and more diverse source data settings. The results are provided in~\cref{tab:scenario2}, where the reference, where the reference setup (GTA5~$\rightarrow$~CS) is marked in gray. We observe that scaling unlabeled target data has little to no, and at times even an adverse, effect on generalization to Cityscapes. Even when adding unlabeled data from the same distribution (Cityscapes-extra), generalization does not improve. Similarly, scaling target data under more diverse source settings does not further benefit UDA. These findings are consistent with recent observations from UDA-Bench~\cite{kalluri2024lagtran}, which also reports limited gains from increasing unlabeled target data. Nevertheless, in all cases, UDA outperforms source-only fine-tuning, and generalization to WildDash2 is maintained or even improved. \textbf{To conclude, scaling unlabeled target data has no significant positive or negative effect on generalization to the target domain.}

\begin{table}[h] 
\centering
\begin{adjustbox}{width=1\linewidth}
\begin{tabular}{lcccc}
\toprule
Setting & Source & Target  & mIoU CS & mIoU WD2 \\ 
\toprule
Source-only FT  & GTA5  & -   & 71.8\phantom{..↑.-0.3}  & 65.5\phantom{..↑.-0.5} \\
\rowcolor{gray!30}
UDA & GTA5  & CS  & 79.8\phantom{..↑.-0.3}  & 69.0\phantom{..↑.-0.5}\\ 
UDA & GTA5 & CS, CS Extra              & 79.5\phantom{.}\reddown~\textcolor{brickred}{-0.3} &  69.4\phantom{.}\greenup~\textcolor{applegreen}{+0.4}\\ 
UDA & GTA5 & CS, BDD, Vistas, DZ, ACDC & 80.3\phantom{.}\greenup~\textcolor{applegreen}{+0.5} & 70.5\phantom{.}\greenup~\textcolor{applegreen}{+1.5}\\ 
\midrule
Source-only FT  & GTA5, S, US  & -& 79.7\phantom{.↑.+1.8}  & 67.3\phantom{.↑.+3.4}\\
\rowcolor{gray!30}
UDA & GTA5, S, US & CS & 81.5\phantom{.↑.+1.8} & 70.7\phantom{.↑.+0.1} \\ 
UDA & GTA5, S, US & CS, CS Extra  & 81.5\phantom{..↑..}\textcolor{gray}{0.0} & 70.8\phantom{.}\greenup~\textcolor{applegreen}{+0.1} \\ 
UDA & GTA5, S, US & CS, BDD, Vistas, DZ, ACDC & 81.3\phantom{.}\reddown~\textcolor{brickred}{-0.2}  & 71.0\phantom{.}\greenup~\textcolor{applegreen}{+0.3} \\ 
\bottomrule
\end{tabular}
\end{adjustbox}
\caption{\label{tab:scenario2} \textbf{Scenario Synth 2A: Scaling target data with small-scale (top) and large-scale (bottom) source data.} Scaling unlabeled real target data in these scenarios has no significant positive or negative effect on generalization to the target domain.} 
\end{table}

We perform another important experiment in which we alter the distribution of the diverse source data. The results, shown in~\cref{tab:scenario2b}, reveal that replacing UrbanSyn~\cite{gomez2023urbansyn} with SynScapes~\cite{Wrenninge2018Synscapes} has a significant adverse effect on source-only fine-tuning, with a drop of 3.8 mIoU. Interestingly, UDA is not significantly affected by this change in datasets. This robustness likely stems from UDA being `regularized’ by its target domain data, making it less sensitive to variations in the source distribution. This is a key finding, as it is one of the few scenarios where UDA shows significant added value, with a gain of 6.3 mIoU over source-only fine-tuning. We come back to this in our discussion. \textbf{To conclude, UDA is less susceptible to distribution changes in the source data than source-only fine-tuning.}

\begin{table}[h] 
\centering
\begin{adjustbox}{width=1\linewidth}
\begin{tabular}{lcccc}
\toprule
Setting & Source & Target  & mIoU CS & mIoU WD2 \\ 
\toprule
\rowcolor{gray!30}
Source-only FT & GTA5, S, SS & -                         & 75.9\phantom{sas+6.3}                      & 67.2\phantom{as+3.7} \\
UDA            & GTA5, S, SS & CS & 80.8~\greenup~\textcolor{applegreen}{+4.9} & 70.0~\greenup~\textcolor{applegreen}{+2.8}\\ 
UDA            & GTA5, S, SS & CS, BDD, Vistas, DZ, ACDC & 82.2~\greenup~\textcolor{applegreen}{+6.3} & 71.3~\greenup~\textcolor{applegreen}{+4.1}\\ 
\midrule
\rowcolor{gray!30}
Source-only FT & GTA5, S, US & -                         & 79.7\phantom{sas+1.3}                      & 67.3\phantom{as+3.7}\\
UDA            & GTA5, S, US & CS & 81.5~\greenup~\textcolor{applegreen}{+1.8} & 70.7~\greenup~\textcolor{applegreen}{+3.4}\\ 

UDA & GTA5, S, US & CS, BDD, Vistas, DZ, ACDC & 81.3\phantom{.}\greenup~\textcolor{applegreen}{+1.6}  & 71.0\phantom{.}\greenup~\textcolor{applegreen}{+3.7} \\ 
\bottomrule
\end{tabular}
\end{adjustbox}
\caption{\label{tab:scenario2b} \textbf{Scenario Synth 2B: Influence of source data composition.} The composition of synthetic source datasets may negatively affect the generalization of source-only fine-tuning and, consequently, raises the added value of UDA.}\
\end{table}

\PAR{Scenario Synth 3: Mixing in few target labels.}
In the third scenario, we use a small amount of labeled real target data. The results in~\cref{tab:synthscenario3} show that UDA can leverage this small amount of labeled images better than source-only fine-tuning. Interestingly, it even performs similarly to a fully-supervised model trained with all Cityscapes labels. Although there are only slight performance differences, these results confirm the trend that synth-to-real UDA systematically outperforms source-only fine-tuning. Academically, these are very strong results, but we reflect on the practical implications in the discussion. \textbf{To conclude, UDA can structurally outperform source-only fine-tuning in synth-to-real scenarios, also when a small amount of labeled target data is available.}  

\begin{table}[h] 
\centering 
\begin{adjustbox}{width=\linewidth}
\begin{tabular}{lccccc} 
\toprule
Setting & Stage 1 & Stage 2 & mIoU CS & WD2 \\
\toprule
\rowcolor{gray!30}
Source-only FT         & GTA5, S, US                             & 1/16 CS                  & 83.0\phantom{.↑.+2.1}                      & 70.0\phantom{.↑.+1.5}  \\
UDA                    & GTA5, S, US~$\rightarrow$~CS            & 1/16 CS~$\rightarrow$~CS & 85.1~\greenup~\textcolor{applegreen}{+2.1} & 71.5~\greenup~\textcolor{applegreen}{+1.5} \\

Fully-supervised       & GTA5, S, US                             & CS                       & 85.1~\greenup~\textcolor{applegreen}{+2.1} & 72.2~\greenup~\textcolor{applegreen}{+2.2}  \\
\bottomrule
\end{tabular}
\end{adjustbox} 
\caption{\label{tab:synthscenario3} \textbf{Scenario Synth 3: Mixing in few target labels.} With minimal real target labels, UDA outperforms source-only fine-tuning and performs similarly to the fully-supervised oracle.} 
\end{table}

\subsection{Real-to-real results}
\label{sec:real_results}

\PARbegin{Scenario Real 1: Scaling source data.}
In the first real-to-real scenario, we measure the added value of UDA over source-only fine-tuning, when more diverse labeled real source data is available. From the results in \cref{tab:scenarioR1}, we can first observe that when less diverse source data is available (top part of the table), UDA is able to significantly improve generalization over source-only fine-tuning. This benefit of UDA on small-scale benchmarks is well-known, but, as described earlier, more diverse source data is arguably a more realistic setting. Considering this more realistic scenario (lower half of the table), we see that UDA has no positive, or even a small negative effect. While the generalization to WildDash2 improves slightly, these results show that UDA is not much of added value when sufficiently rich source data is available. \textbf{To conclude, when real source data is diverse enough, UDA has little added value over source-only fine-tuning}.

\begin{table}[h] 
\centering 
\begin{adjustbox}{width=1\linewidth}
\begin{tabular}{lccccc} 
\toprule
Setting & Source & Target  & mIoU CS & mIoU WD2 \\
\midrule
\rowcolor{gray!30}
Source-only FT  & BDD               & -   & 77.7\phantom{sas+?.?}                                      & 67.9\phantom{sas+?.?} \\
UDA             & BDD               & CS  & 80.3~\greenup~\textcolor{applegreen}{+2.6}                 & 70.4~\greenup~\textcolor{applegreen}{+3.8} \\
\midrule
\rowcolor{gray!30}
Source-only FT  & BDD, Vistas, ACDC & -   & 83.0\phantom{sas-0.9}                                      & 73.9\phantom{sas-0.4} \\
UDA             & BDD, Vistas, ACDC & CS  & 82.7~\reddown~\textcolor{brickred}{-0.3}                   & 74.7~\greenup~\textcolor{applegreen}{+0.8}\\
\bottomrule
\end{tabular}
\end{adjustbox} 
\caption{\label{tab:scenarioR1} \textbf{Scenario Real 1: Scaling source data.} When including more labeled real source datasets, UDA has little added value over source-only fine-tuning. 
} 
\end{table}

\PAR{Scenario Real 2: Mixing in few target labels.}
In the second real-to-real scenario, we use a small amount of labeled target data. Contrary to the previous experiment, we now see a positive effect of UDA over source-only fine-tuning of VFMs in~\cref{tab:scenario3}. UDA is able to obtain the same performance as the fully-supervised model by only using 1/16th of its labels. But again, UDA's relatively modest difference with source-only fine-tuning might not justify its real-world application. \textbf{To conclude, UDA offers some benefit when a small amount of target labels are available, but its advantage remains limited when sufficiently diverse real source data is available.}
 
\begin{table}[h] 
\centering 
\begin{adjustbox}{width=\linewidth}
\begin{tabular}{lccccc} 
\toprule
Setting & Stage 1 & Stage 2 & mIoU CS & WD2 \\
\midrule
\rowcolor{gray!30}
Source-only FT                & BDD, Vistas, ACDC                            & 1/16 CS                  & 83.5\phantom{.↑.+1.2} & 71.7\phantom{.↑.+2.1} \\

UDA                           & BDD, Vistas, ACDC~$\rightarrow$~CS           & 1/16 CS~$\rightarrow$~CS & 84.7~\greenup~\textcolor{applegreen}{+1.2} & 73.8~\greenup~\textcolor{applegreen}{+2.1} \\ 
Fully-supervised              & BDD, Vistas, ACDC                            & CS                       & 84.7~\greenup~\textcolor{applegreen}{+1.2} & 73.5~\greenup~\textcolor{applegreen}{+1.8}   \\
\bottomrule
\end{tabular}
\end{adjustbox} 
\caption{\label{tab:scenario3} \textbf{Scenario Real 2: Mixing in few target labels.} With a small amount of labeled target data, UDA achieves performance comparable to fully-supervised learning with all target labels.}
\end{table}

\section{Discussion}
\label{sec:discussion}
UDA for synth-to-real and in scenarios with few target labels consistently outperforms source-only fine-tuning, achieving generalization close to fully-supervised learning. With only 1/16 of the Cityscapes labels, it even matches fully-supervised performance across all evaluated datasets. While these results are impressive from an academic viewpoint, the actual added value over source-only fine-tuning remains modest, typically only a few mIoU points. This raises the question of whether UDA’s improvements justify its added training complexity, particularly as VFMs continue to improve and source-only fine-tuning becomes increasingly effective. We dare to doubt if these results alone are a sufficient reason to warrant the real-world usage of UDA. Although niche applications exist, \ie, scenarios with limited labeled source data with a substantial domain gap to the target, we argue that such conditions are rare in mainstream autonomous driving. Therefore, we don't see UDA as a key enabler for autonomous driving, because more straightforward source-only fine-tuning of VFMs can achieve practically similar results.

This, however, assumes that labeled source data is sufficiently diverse to cover all target domains, an assumption that may be difficult to guarantee. In cases where domain gaps are substantial and labeled target data is unavailable, UDA may still serve as a fallback mechanism by adapting the model without requiring new annotations. Our findings suggest that, in such cases, leveraging UDA with all available labeled source data and unlabeled target data yields a model that generalizes better than source-only fine-tuning, approaching fully-supervised performance. 
Future research should explore whether this approach remains effective across a broader range of domains and dynamic environmental conditions. Ultimately, our study challenges the assumption that UDA is broadly necessary for autonomous driving but provides a refined perspective on when and where it retains value, not as a standard training paradigm, but as a targeted adaptation strategy when domain generalization is a critical concern.

\section{Conclusion}
\label{sec:conclusions}
To conclude, we evaluated the added value of UDA in the era of VFMs, across both synth-to-real and real-to-real scenarios. While UDA achieves impressive generalization, even matching fully-supervised performance with only a small amount of labeled target data, its added value over source-only fine-tuning of VFMs is limited. UDA improves synth-to-real performance, but its benefits diminish as the diversity of synthetic source data increases. Scaling unlabeled target data has no meaningful impact on generalization to the target domain. However, UDA is less sensitive to distribution changes in the source data than source-only fine-tuning. In real-to-real scenarios, when source data is sufficiently diverse, UDA offers little advantage. These findings challenge the practicality of UDA for autonomous driving. Still, in cases where domain gaps remain substantial and labeled target data is unavailable, UDA may serve as a fallback to adapt models without requiring new annotations. By systematically evaluating UDA in the VFM era, our work offers a new perspective on its role in perception for autonomous driving. Our findings refine the understanding of when and where UDA is beneficial, paving the way for more effective adaptation strategies in future autonomous systems.

\section*{Acknowledgements}

The project EdgeAI “Edge AI Technologies for Optimised Performance Embedded Processing” is supported by the Chips Joint Undertaking and its members including top-up funding by Austria, Belgium, France, Greece, Italy, Latvia, Netherlands, and Norway under grant agreement No 101097300.

The project Ecomobility has received funding within the Chips Joint Undertaking (Chips JU), a public-private partnership in collaboration with the Horizon Europe (HORIZON) Framework Programme and National Authorities under grant agreement number 101112306.

This work made use of the Dutch national e-infrastructure with the support of the SURF Cooperative, using grant no. EINF-9321 and EINF-12906, which is financed by the Dutch Research Council (NWO).

\section*{Disclaimer}
Funded by the European Union. Views and opinions expressed are however those of the author(s) only and do not necessarily reflect those of the European Union or the Chips Joint Undertaking. Neither the European Union nor the granting authority can be held responsible for them.

{
    \small
    \bibliographystyle{ieeenat_fullname}
    \bibliography{main}
}

\end{document}